# Assessing Pedestrian Behavior Around Autonomous Cleaning Robots in Public Spaces: Findings from a Field Observation

Maren Raab[1], Linda Miller[1], Zhe Zeng[1], Pascal Jansen[2], Martin Baumann[1] and Johannes Kraus[3]

*Abstract*— **As autonomous robots become more common in public spaces, spontaneous encounters with laypersons are more frequent. For this, robots need to be equipped with communication strategies that enhance momentary transparency and reduce the probability of critical situations. Adapting these robotic strategies requires consideration of robot movements, environmental conditions, and user characteristics and states. While numerous studies have investigated the impact of distraction on pedestrians' movement behavior [1-4], limited research has examined this behavior in the presence of autonomous robots. This research addresses the impact of robot type and robot movement pattern on distracted and undistracted pedestrians' movement behavior. In a field setting, unaware pedestrians were videotaped while moving past two working, autonomous cleaning robots. Out of *N* = 498 observed pedestrians, approximately 8% were distracted by smartphones. Distracted and undistracted pedestrians did not exhibit significant differences in their movement behaviors around the robots. Instead, both the larger sweeping robot and the off-set rectangular movement pattern significantly increased the number of lateral adaptations compared to the smaller cleaning robot and the circular movement pattern. The off-set rectangular movement pattern also led to significantly more close lateral adaptations. Depending on the robot type, the movement patterns led to differences in the distances of lateral adaptations. The study provides initial insights into pedestrian movement behavior around an autonomous cleaning robot in public spaces, contributing to the growing field HRI research.**

## I. Introduction

Due to major advancements in artificial intelligence and automation technology, autonomous robots are becoming increasingly common in public spaces [5]. Public spaces are usually characterized by their crowdedness, unpredictability, and complexity [6]. In such surroundings, robotic tasks need to be coordinated with coincidentally present pedestrians who are not directly involved in the task. From a 1:1 interaction perspective in cooperative human-robot interaction (HRI), the requirements for robotic interaction strategies and communication change drastically. Instead of high levels of coordination and feedback with human users, the paradigm changes to unobtrusiveness and communication by exception in public spaces. Especially in the face of the unpreparedness of surrounding people, considerations of safety become even more relevant addressing, especially user states that, from a Human Factors perspective, reduce attentiveness and readiness to effectively coordinate one's path with the robot. Distraction is a key factor in safety-critical human-technology interactions but has been largely overlooked in public HRI. Simultaneously, distraction while moving through public spaces might be an increasing challenge as reflected in the rise of the proportion of mobile internet users in Germany in the years 2015 to 2022 increased from 54% to 84% [7]. In Germany (Jan-Aug 2023), driver distraction caused 4,941 injury accidents vs. 271,349 from other human errors [8]. Considering this, it is of little surprise that, for example, adolescents frequently report using their smartphones while walking in public spaces [9]. Termed *smombie* ("smartphone zombie") [10] behavior, this phenomenon poses a substantial risk to the safety of the distracted pedestrian and their surroundings. Because of these simultaneous developments of increasing complexity of public spaces and *smombie* behavior, understanding the dynamic in pedestrians' and autonomous robots' encounters is essential. Whilst prior studies have investigated the motion changes in pedestrians through (smartphone-) distraction [1-4], the literature on distracted pedestrians around autonomous robots is still scarce. There has been a gap in the literature evaluating service robots' behavior when they are not directly interacting with their passersby, as is the case with cleaning robots [11]. The present study aims to provide insights into how distracted and undistracted pedestrians react differently to autonomous service robots in the wild. Over this, the role of robot type (mainly reflected in robot size) and robotic movement patterns for behavioral changes are investigated in a video-based observational study in a realistic field setting with autonomous robots. 'Autonomous' here refers to robots capable of independent navigation and decision-making without direct human control [12]. Our research specifically examines service robots, meaning robots that perform a designated task [12], in our case a cleaning robot in a public space. These robots must perform their assigned tasks efficiently while navigating freely through the environment, which directly influences pedestrian movement behavior [13,14]. Their unconstrained mobility presents unique challenges in HRI within crowded and unpredictable settings.

## II. Related Work

### A. Human-Robot Encounters in Public Spaces

Early research on HRI mainly focused on exploring direct dyadic interactions between robots and humans in a shared

[1]MR, LM, ZZ, and MB are with the Human Factors Department, Ulm University 89081, Germany (corresponding author: zhe.zeng@uni-ulm.de).

[2]PJ is with the Institute of Media Informatics, Ulm University 89081, Germany.

[3]JK is with the Department of General Experimental Psychology, Johannes-Gutenberg University of Mainz, 55122, Germany.

*Research supported by the BMBF ZEN-MRI project.

task setting [15]. However, with more coincidental encounters taking place due to the increasing presence of autonomous robots in public spaces, spontaneous human-robot encounters (HRE) have moved into the center of interest for researchers. These encounters coined the term *InCoP*s, referring to "incidentally copresent persons" [16], pedestrians who simply "happen to be there" [16]. [17] argue that studies should focus more on exploring implicit interactions taking place between these *InCoP*s and autonomous robots, i.e. through analyzing video recordings.

Several studies have provided insights into different reactions of people to robots in public spaces [18-20]. One field study, conducted by [18] at a train station, researched the passerby' reactions and conflicts to an autonomous cleaning robot also employed in this study. Their video data analysis showed that the majority of the pedestrians noticed the robot, with about one-third ignoring it. One in four pedestrians actively evaded the robot, meaning they adapted either their walking pace or their trajectory. Similarly, [21] explored *InCoP*s' reactions to a seemingly autonomous delivery robot operating on a sidewalk through a Wizard-of-Oz setup. He found that most of the encountered pedestrians showed neutral to positive reactions to the robot. However, some pedestrians reported that they struggled to predict how the robot would act and expressed concerns about it being a danger for children, who might be inattentive at times. The study by [17] on *InCoP*s' reactions to an autonomous delivery robot on public streets showed that both the robots and people on the sidewalks must navigate around each other, resulting in the robot, the pedestrian, or both making adaptations to their trajectories.

This study expands the existing literature on HRE by exploring natural encounters between pedestrians and two autonomous cleaning robots in a public space under realistic conditions. Whilst prior studies have focused on specific groups of *InCoP*s, such as pedestrians and bicyclists [22], there is a lack of studies looking into how distracted pedestrians may differ from undistracted pedestrians around autonomous robots.

### B. Impact of Distraction on Movement Behaviors

The extent to which distracted driving is problematic has been rigorously studied. One of the main shortcomings of distracted driving is inattentional blindness - the inability to recognize new stimuli in the environment [4,23]. Studies have shown that pedestrians, like drivers, often engage in multitasking during walking [24]. [25] observed that almost 30% of the pedestrians of their sample in a Spanish city were using their smartphones in some way while walking.

As distracted walking is common, researchers have been studying the effects of distraction on pedestrians. [4] compared pedestrians displaying different types of distraction – talking on a cell phone, listening to music, walking in pairs, and walking without any device as an undistracted baseline – regarding their movement and their awareness of other people in the environment, as well as how likely they were to notice a clown on a unicycle. They found that pedestrians who were engaged in using their cell phones while walking were the least likely to notice the clown and thus most likely to display inattentional blindness, whereas 71% of pedestrians walking in pairs were aware of the clown. Additionally, those pedestrians using their cell phones were slower than others, altered directions, weaved more, and were less aware of other pedestrians in their surroundings. [2] explored how smartphone use influences pedestrians' situational awareness to roadside events and arrived at similar conclusions in a semi-virtual environment. In a dual task paradigm, a significant increase in perceived workload was shown, as well as reduced awareness of roadside events.

Besides the effects on inattentional blindness and awareness of roadside events, studies have shown that distraction whilst walking has an impact on the gait, as well as movement behavior of the pedestrian. [3] investigated the effects of walking during texting on the ability of pedestrians to effectively navigate obstacles in daily life, such as curbs, steps, bollards placed on the street, and other pedestrians. Pedestrians involved in either texting or solving a mathematics exercise took significantly longer to complete the obstacle course compared to a control group and had different gait patterns. Against their hypothesis, [3] did not find an increased risk of tripping in the distracted groups. They concluded that pedestrians involved in texting while walking adopt a protective gait that helps them prevent potential accidents. Although their study found no increased lateral deviations for distracted pedestrians avoiding obstacles, research by [26] and [27] contradicts this, showing distracted pedestrians make stronger lateral deviations from a straight path compared to undistracted pedestrians.

### C. Movement Behavior of Pedestrians around Robots in Public Spaces

Considering the aforementioned increase of robots being deployed in public spaces, researchers have started to investigate the effects robots have on pedestrian movement behavior and crowd dynamics. [13] conducted a study aiming at a better understanding of the differential effects of the presence of different robot types in public space. Their data indicates that pedestrians are on average significantly slower, as well as accelerating less in the presence of a robot compared to no robot being present. Using Hall's theory of proxemics [28] system to classify proximity to the robot, they also analyzed whether the pedestrians' speed was varying depending on how close they were to the robot. Results show that within the same scenario, pedestrians with less than 2.1m between them and the robot - and thus within what [28] would consider close social space - were slower than those with more space to the robot. Additionally, those in close proximity to the robot were found to have a slightly reduced path efficiency in two of three scenarios. A study from a shopping mall in Japan aimed at calibrating and validating a mathematical model describing the movement of pedestrians around a social robot in public spaces [14]. They found that pedestrians tend to respond to a service robot more noticeably and from a greater distance compared to another human.

In comparing the effect of different robots on human evaluation and reaction, it is a challenge to clearly trace back the source of resulting differences, as different robot types normally differ in more than one aspect. One of the most researched variables in this respect is robot size. [29] investigated the psychological effects of different robot heights. Results indicate that participants prefer to keep a larger distance towards bigger robots. Additionally, participants' posture influenced their comfort. They felt most threatened by the largest robot, no matter their posture. For the smallest robot, no difference for the postures was found, whereas for the standard-size robot, participants felt more threatened when they were seated. Similarly, [30] compared two robot heights and two positions of the robot's manipulator. They found that a shorter robot was associated with lower levels of anxiety and tolerated to come closer than a taller robot.

For humans to effectively adapt their motion behavior when encountering a robot, they must understand the robot's intentions [31]. In public spaces, pedestrians need information about a robot's motion intent to prevent collisions and to feel safe [32]. A robot's intention can be communicated implicitly (i.e. by speed and movement) and explicitly (i.e. by projecting the robot's planned trajectory) [6]. In a study [33], participants were exposed to three robot navigation strategies, including two autonomous ones. The strategies used were *ORCA* (Optimal Reciprocal Collision Avoidance), which intends to generate fluid, accident-free and natural-appearing simulations of multi-agent scenarios, using pre-defined collision-avoidance rules, *Social Momentum* (SM), which emphasizes understanding the intentions of others and integrating this awareness into its trajectory planning, adapting its movements based on the observed intentions of the other agent, and lastly, the *Teleoperation Strategy* (TE), in which a human teleoperator intents to generate natural interactions. Their analysis concludes that pedestrians' acceleration is lower when moving in the realms of an autonomous robot compared to a teleoperated robot. Participants' movement was found to be smoother and followed more consistent paths when navigating in the *Social Momentum* condition compared to the teleoperated robot. This demonstrates that a robot's behavioral design significantly influences the walking behavior of nearby pedestrians.

### D. Hypotheses and Research Question

**H1:** *Compared to undistracted pedestrians, distracted pedestrians exhibit more frequent adjustments of their movement behavior (H1a) and later adjustments (i.e., closer to the robot, H1b).*

**H2:** *The larger sweeping robot, compared to the smaller cleaning robot, elicits earlier and more frequent movement adjustments from all pedestrians (H2a), with distracted pedestrians adjusting their movement behavior more frequently (H2b) but undistracted pedestrians adjusting earlier (H2c).*

**RQ1:** *What impact does the robot's movement pattern (circular vs. off-set rectangular) have on the movement behavior of distracted and undistracted pedestrians?*

## III. Method

### A. Autonomous Cleaning Robots

Two autonomous cleaning robots were used in this study (see Figure 1). The SR1300 (116.3x100.3x150.6cm) was originally designed for large halls and warehouses, moving at 0.03-1.11m/s. The CR700 (100.0x80.5x98.0cm) is suitable for deployment in various facilities, moving at 0.3-0.8m/s [34]. Both operate autonomously, avoiding collisions with obstacles by either stopping or moving around them.

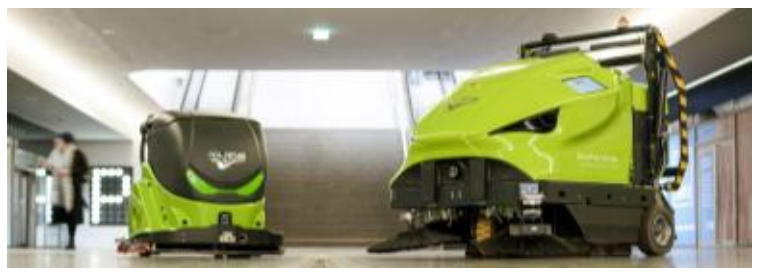

Figure 1. Cleaning robots from ADLATU: CR700 (left) and SR1300 (right).

### B. Study Set-Up

A two-day observational video study was conducted in August 2023 in an underground passage in Ulm, Germany. Only pedestrians who walked through the entire robot-occupied area were included. Pedestrians who left the area at some point to take one of the side exits were not considered. The two robots were used alternately in sets of about 60 minutes each, covering half the space in a *circular* and an *off-set rectangular movement pattern*, respectively (see Figure 2).

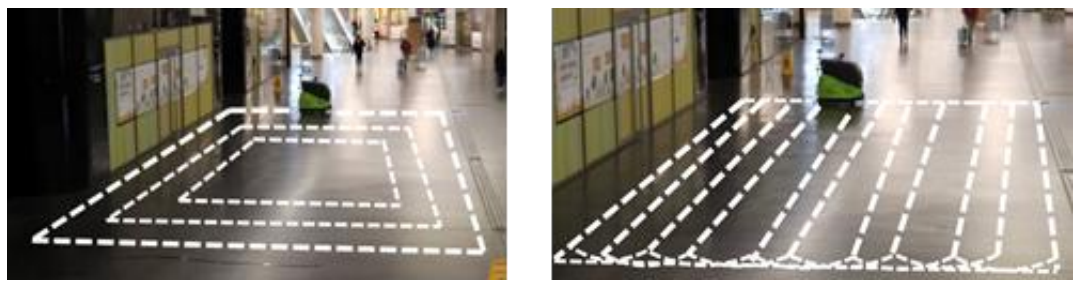

Figure 2. Observation of space and robot movement patterns. The white lines indicate the robots' movement patterns. The circular movement pattern (left); the off-set rectangular movement pattern (right).

Four static cameras and three robot-mounted cameras, which allowed a 360° view of the observed space, were used for the recordings (see Figure 3). Pedestrians were informed of filming via display stands, and experimenters were present to intervene in case of problems. The filming and use of data were carefully coordinated with legal experts for data privacy to be in full compliance with applicable data protection regulations. About two hours of video material were analyzed for this study. The selected video clips included an equal distribution of video sequences across both robot types and movement patterns, with one clip of 30 minutes per robot and movement pattern each.

Overall, $N$ = 498 pedestrians were coded. As a first step in the coding procedure, they were categorized into six different age groups and assigned a perceived gender based on visual assessment of the video footage. Here, the limitations of these appearance-related categorizations must be noted. The majority of pedestrians were estimated to be between the age of 21 - 45 years (43%), followed by the age group 46 - 64 years (27%). 54% of the coded passersby were classified as female, the other 46% as male. Data were collected from $n$ = 276 pedestrians for the cleaning robot and $n$ = 222 pedestrians for the sweeping robot. To summarize, $n$ = 257 (56%) undistracted pedestrians encountered the cleaning robot and $n$ = 203 (44%) the sweeping robot, compared to $n$ = 19 (50%) distracted pedestrians for the cleaning robot and sweeping robot each. For the cleaning robot, $n$ = 146 (53%) encountered the robot moving in the circular pattern and $n$ = 130 (47%) pedestrians moving in the off-set rectangular moment pattern. For the sweeping robot, $n$ = 101 (45%) encountered it following the circular movement pattern and $n$ = 121 (55%) encountered it following the off-set rectangular movement pattern.

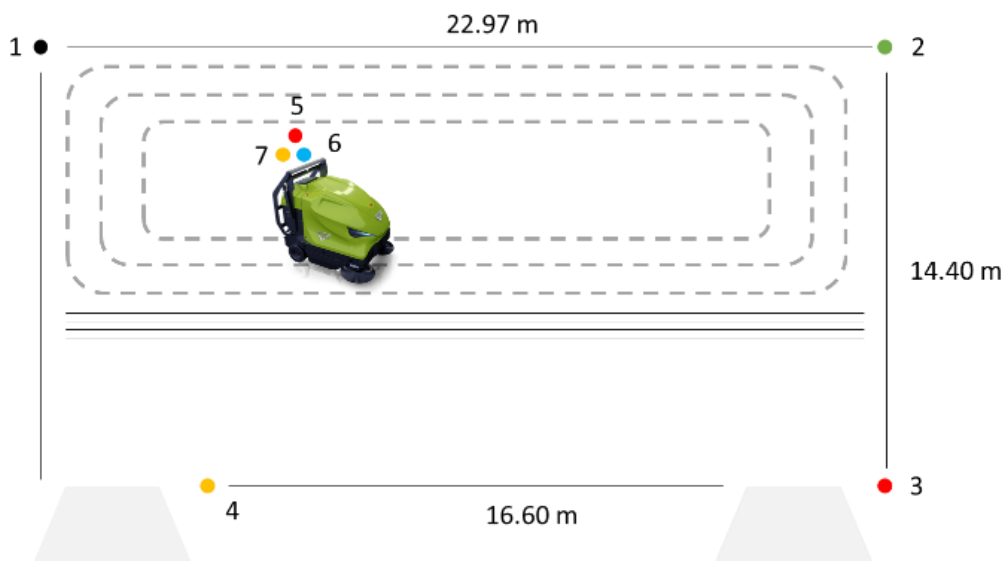

Figure 3. Cameras 1-4 are located at the four corners of the venue, and cameras 5-7 are fixed on the robot to record a 360° field of view.

### C. *Variables and Analysis Procedure*

The study assessed the following variables: distraction, awareness, trajectory adaptations and their distances. Distraction was defined as visual engagement with a smartphone, such that pedestrians observed *looking at a phone* at any point during the encounter were classified as "*distracted*"; all others were considered "*undistracted*". Awareness was classified as pedestrians who did or did not look at the robot. We analyzed whether the pedestrian adapted their trajectory upon encountering the robot. For this, four intensities of adaptations were defined: No adaptation, slight adaptation (< 45° angle change in walking trajectory), medium adaptation (45° - 89°) and strong adaptation (≥ 90°). The distances at which pedestrians initiated their adaptation was classified in accordance with Hall's theory of proxemics [26] system: Close (< 1.2m) and far adaptations (> 1.2m) relative to the robot. The video material was coded by one and a second coder independently coded 12% of the data to estimate interrater- reliability. The Cohen's Kappa values indicate varying agreement levels: substantial for age (.634, $p$ < .001), fair for awareness (.263, $p$ = .006), trajectory adaptations (.291, $p$ < .001), and distance of adaptations (.28, $p$ < .001), while distraction shows slight agreement (.133, $p$ = .005). Significant *p-values* suggest the agreement is not due to chance.

To analyze differences in movement behavior between distracted and undistracted pedestrians across robot types and movement patterns, two binomial logistic regressions were calculated. The first used trajectory adaptations as the dependent variable, while the second used adaptation distance. Both regressions included distraction, robot type, movement pattern, and their interactions as predictors. Trajectory adaptations and adaption distance were dummy-coded, with no adaptation and far adaptation as references. Pedestrians who did not display lateral adaptations were excluded from distance-related calculations.

## IV. Results

### A. *Awareness and Distraction*

Of the $N$ = 498 coded passersby, $n$ = 452 (91%) were looking at the robot, while $n$ = 46 (9%) were not looking at the robot. Overall, $n$ = 38 (8%) pedestrians were classified as distracted, as they were looking at their phone at least once during the encounter with the robot. $n$ = 11 (29%) of the distracted pedestrians did not look at the robot, as opposed to $n$ = 35 (8%) of the undistracted pedestrians. $n$ = 23 (61%) of the distracted pedestrians were between the age of 20 - 45, meaning 11% of pedestrians in this age group were distracted, followed by $n$ = 8 (21%) in the age group of 12 - 19 years, where the distraction was proportionally the highest with 12%.

### B. *Movement Behavior of Distracted Pedestrians compared to Undistracted Pedestrians*

In H1, it was assumed that distracted pedestrians exhibit more frequent (H1a) and closer (H1b) adjustments of their movement behavior in reaction to the robots than undistracted pedestrians. Table 1 displays the number of pedestrians for each intensity of lateral adaptations, for both undistracted and distracted pedestrians, as well as the distances at which they were initiated. 45% of the undistracted and 50% of the distracted pedestrians did not adapt their trajectory upon encountering the robot. The majority of pedestrians-initiated adaptations far from the robot (> 1.2m), including 96% of undistracted and 89% of distracted pedestrians. The logistic regressions (see Table 2 and Table 3) indicated that there was no main effect of distraction for the lateral adaptations, nor the distance of lateral adaptations. Therefore, H1 must be rejected.

TABLE 1. Number Of Pedestrians For Adaptation Intensity And Distances Between Undistracted And Distracted Pedestrians

| *Adaptation Type* | *Undistracted* | *Distracted* |
|---|---|---|
| *Intensity of lateral adaptations* | | |
| No | 206 (45%) | 19 (50%) |
| Slight | 156 (34%) | 12 (32%) |
| Medium | 85 (18%) | 4 (11%) |
| Strong | 13 (3%) | 3 (8%) |
| *Distance of lateral adaptations* | | |
| Close | 11 (4%) | 2 (11%) |
| Far | 243 (96%) | 17 (89%) |

### C. *Impact of Robot Type*

H2a postulated that the larger sweeping robot would prompt more frequent and earlier adjustments compared to the smaller cleaning robot from both pedestrian groups (H2a), with distracted pedestrians adjusting more frequently (H2b), but undistracted pedestrians adjusting earlier (H2c). Overall, 52% did not make lateral adaptations for the cleaning robot and 36% for the sweeping robot. Adaptations were mostly initiated at a far distance (cleaning robot: 93%, sweeping robot: 97%). Partially supporting H2a, the robot type significantly affected the frequency of lateral adaptations (more for the larger sweeping robot), but not the distance at which they occurred. Table 4 summarizes the lateral adaptations and related distances for both robot types by the distraction groups. However, no significant interaction between distraction and the robot type occurred for neither lateral adaptions nor the distance, contradicting H2b and H2c.

TABLE 2. BINOMIAL LOGISTIC REGRESSION RESULTS FOR INTENSITY OF LATERAL ADAPTATIONS

| Predictor | b | SE | p | OR |
|---|---|---|---|---|
| Intercept | 0.40 | 0.04 | <.001*** | 1.50 |
| Undistracted vs. distracted | 0.10 | 0.16 | .55 | 1.10 |
| Circular vs. Off-set rectangular | 0.17 | 0.06 | < .01** | 1.18 |
| Cleaning vs. Sweeping robot | 0.18 | 0.07 | <.01** | 1.20 |
| Dis [a] * MovPa [b] | -0.33 | 0.23 | .15 | 0.72 |
| Dis * RT [c] | -0.08 | 0.28 | .78 | 0.92 |
| MovPa * RT | -0.06 | 0.09 | .53 | 0.94 |
| Dis * MovPa * RT | 0.20 | 0.35 | .58 | 1.22 |

[a] Distraction, [b] Movement Pattern, [c] Robot Type

TABLE 3. BINOMIAL LOGISTIC REGRESSION RESULTS FOR DISTANCE OF LATERAL ADAPTATIONS

| Predictor | b | SE | p | OR |
|---|---|---|---|---|
| Intercept | 0.02 | 0.03 | .52 | 1.02 |
| Undistracted vs. distracted | -0.02 | 0.10 | .85 | 0.98 |
| Circular vs. Off-set rectangular | 0.08 | 0.04 | <.05* | 1.09 |
| Cleaning vs. Sweeping robot | 0.04 | 0.04 | .37 | 1.04 |
| Dis * MovPa | 0.25 | 0.16 | .11 | 1.28 |
| Dis * RT | -0.04 | 0.16 | .82 | 0.97 |
| MovPa * RT | -0.14 | 0.05 | <.05* | 0.87 |
| Dis * MovPa * RT | -0.07 | 0.22 | .74 | 0.93 |

## D. *Impact of Robot Movement*

Lastly, the impact of the robots' movement pattern on the movement behavior of distracted and undistracted pedestrians was explored in RQ1. Less adaptations occurred for the circular movement pattern, with 52% pedestrians not changing their trajectory compared to 39% for the off-set rectangular movement pattern that the robots followed. Table 4 shows the number of each lateral adaptation type for both movement patterns, as well as the distances of lateral adaptations for both distraction groups. Regarding the distances at which lateral adaptations were initiated, only 4% undistracted pedestrians and 0% distracted pedestrians-initiated adaptations within 1.2 meters of the robot for the circular movement pattern. In contrast, for the off-set rectangular movement pattern, 5% undistracted pedestrians and 18% distracted pedestrians initiated their adaptations within 1.2 meters of the robot. The results show a significant main effect on both dependent variables, increasing the odds of a lateral adaptation and close adaptations occurring for the off-set rectangular movement pattern of the robot. In addition, a significant interaction between movement pattern and robot type was found, indicating that movement pattern and RT do not act independently; their combined effect plays a meaningful role in predicting the probability of the outcome related to the distance of lateral adaptations.

TABLE 4. NUMBER OF PEDESTRIANS FOR EACH ADAPTATION INTENSITY AND THEIR DISTANCES BETWEEN UNDISTRACTED AND DISTRACTED PEDESTRIANS FOR BOTH ROBOT TYPES AND MOVEMENT PATTERNS

| | ***Undis-tracted*** | ***Distrac ted*** | ***Sum*** | ***Undis-tracted*** | ***Distract -ed*** | ***Sum*** |
|---|---|---|---|---|---|---|
| | ***Cleaning robot*** | | | ***Sweeping robot*** | | |
| *Intensity of lateral adaptations* | | | | | | |
| No | 133/52% | 11/58% | 144/52% | 73/36% | 8/42% | 81/36% |
| Slight | 74/29% | 6/32% | 80/29% | 82/40% | 6/32% | 88/40% |
| Medium | 39/15% | 2/11% | 41/15% | 46/23% | 2/11% | 48/22% |
| Strong | 11/4% | 0/0% | 11/4% | 2/1% | 3/16% | 5/2% |
| *Distance of lateral adaptations* | | | | | | |
| Close | 8/7% | 1/13% | 9/7% | 3/2% | 1/9% | 4/3% |
| Far | 116/94% | 7/88% | 123/93% | 127/98% | 10/91% | 137/97% |
| | ***Circular*** | | | ***Off-set*** | | |
| *Intensity of lateral adaptations* | | | | | | |
| No | 121/52% | 7/47% | 128/52% | 85/37% | 12/52% | 97/39% |
| Slight | 59/25% | 6/40% | 65/26% | 97/43% | 6/26% | 103/41% |
| Medium | 45/19% | 2/13% | 47/19% | 40/18% | 2/9% | 42/17% |
| Strong | 7/3% | 0/0% | 7/3% | 6/3% | 3/13% | 9/4% |
| *Distances of lateral adaptations* | | | | | | |
| Close | 4/4% | 0/0% | 4/3% | 7/5% | 2/18% | 9/6% |
| Far | 107/96% | 8/100% | 115/97% | 136/95% | 9/82% | 145/94% |

# V. DISCUSSION

## A. *Sharing Spaces with Robots in Public*

In this study, 8% of pedestrians were classified as distracted by their phone, a lower proportion than reported in prior research [1,25]. This discrepancy may be attributed to methodological differences; while [25] employed direct observations, our study relied on video data analysis. The majority of pedestrians (91%) were observed looking at the robot, with only 9% classified as not looking. Notably, even those pedestrians not looking at the robot demonstrated an apparent awareness of its presence through their behavioral adaptations and successful navigation around it. This suggests that visual attention alone may not fully capture pedestrians' awareness of and response to robots. Distraction appeared to influence whether pedestrians looked at the robot or not, with 29% of distracted pedestrians not looking at the robot, compared to only 8% of the undistracted group. Although the number of distracted pedestrians looking at the robot is higher than in previous research by [4] on inattentional blindness, where 75% of cell phone users did not notice a clown on a unicycle, distraction still seems to influence pedestrians' perception of the robot. Nevertheless, the high percentage of pedestrians looking at the robot suggests that the robot attracts attention and is highly noticeable in the environment.

Overall, a substantial proportion of pedestrians (55% undistracted, 50% distracted) changed their trajectory upon encountering the robot, which is higher compared to prior research such as [18], where only about one-fourth of pedestrians evaded the robot. Our findings also contrast with several previous studies [18,35,36] that reported ignoring or disinterest as the most common pedestrian responses to robots. However, direct comparison with previous research requires careful consideration due to significant methodological differences. For instance, [35] utilized a simulated environment, which may not fully capture the complexities of real-word interactions. [36] employed a humanoid robot substantially smaller than our robots. Compared to [18], our study environment offered more space for robot avoidance, potentially facilitating a higher frequency of adaptive behaviors. Moreover, our coding approach focused on observable behavioral adaptations rather than inferring interest or engagement. Thus, pedestrians we coded as making lateral adjustments might have been categorized as 'ignoring' or 'uninterested' in other studies. This distinction highlights a key aspect of HRI in public spaces: even when pedestrians appear uninterested, the robot's presence may still influence their movement patterns. These observations underscore the importance of clearly defining and operationalizing behavioral measures in HRI studies that extend beyond initial curiosity and engagement behaviors. Implementing objective, quantifiable measures of pedestrian behavior, such as precise movement tracking, alongside subjective assessments could provide a more comprehensive understanding of human responses to robots in public spaces.

### B. *Movement Behavior of Distracted Pedestrians compared to Undistracted Pedestrians*

H1 investigated the differences in movement behavior between distracted and undistracted pedestrians when encountering robots in public spaces. The analysis revealed no significant differences in movement behavior (i.e. lateral adaptations and related distances) between distracted and undistracted pedestrians when encountering robots in public spaces. On a descriptive level, data shows that strong adaptations were proportionally more frequent among distracted pedestrians. This pattern aligns with findings from crossing behavior studies [37,38], suggesting that distracted pedestrians may initiate adaptations later, potentially necessitating stronger compensatory movements to avoid collisions. In line with this argumentation, distracted pedestrians indeed showed more close adaptations compared to undistracted ones (11% and 4% respectively descriptively). However, given the limited number of distracted pedestrians, these observations should be interpreted cautiously. The results indicate that even distracted pedestrians had sufficient information to make adaptations in time. Notably, the 'distracted' group included individuals who looked at their phone before getting close to the robot but may have disengaged from the device upon noticing the robot or periodically look up from their device. In addition, distracted pedestrians might still have been using their peripheral vision to navigate around obstacles. Further, the robots might have been well noticeable through their movement and sound, allowing even distracted pedestrians to adjust their behavior similarly to undistracted ones. Results may differ with robots that introduce a higher threat level or with a broader conceptualization of distraction.

### C. *Impact of Robot Type*

H2 predicted that the larger sweeping robot, compared to the smaller cleaning robot, would elicit earlier and more frequent movement adjustments from pedestrians. It also hypothesized differences between distracted and undistracted pedestrians regarding the timing and frequency of these adjustments. In support of H2a, significantly more lateral adaptations were found when pedestrians encountered the sweeping robot compared to the cleaning robot. This aligns with prior studies [29,30] which found that pedestrians preferred more space around larger robots compared to smaller ones. Specifically, the presence of the sweeping robot increased the odds of a trajectory adaptation 1.2 times. This difference may be attributed to several factors. First, the larger size of the sweeping robot likely made it more challenging to navigate around for both distracted and undistracted pedestrians. Second, the sweeping robot's appearance, including its size, sound, visible brushes, and movement behavior, might have elicited increased anxiety or threat in pedestrians, potentially leading to a stronger avoidance response. Third, the sweeping robot occupied more of the observation space than the cleaning robot, inherently necessitating more frequent adaptations from pedestrians. Given these multiple variables, we cannot attribute the increased adaptations solely to one of those factors. Future research should aim to isolate these variables to determine their individual impacts on pedestrian movement behavior. Although significant differences were found for the robot types, the undistracted and distracted pedestrians did not differ regarding the lateral adaptations and the distances based on the robot type. Thus, H2b and H2c were rejected. As argued before, distracted pedestrians might still have perceived both robots through their peripheral vision and the robots' noises. As described by [39], when visual cues are unavailable (as in the case of distraction by a phone), pedestrians may rely on auditory cues to locate objects. This enable distracted pedestrians to make timely adaptations based on the robot's sound, potentially negating differences with undistracted pedestrians.

### D. *Impact of Robot Movement*

RQ1 explored the impact of robot movement patterns on the movement behavior of distracted and undistracted pedestrians. Our findings revealed that pedestrians were more likely to adapt their trajectory when encountering robots following an off-set rectangular pattern compared to circular movement. Similar to the effect of robot type, the off-set rectangular pattern occupies more space and involves more frequent and narrow turns, increasing the likelihood of lateral adaptations. Previous research [40] found that seated participants preferred robots approaching from the front left or right, rather than directly frontal. [6] also suggested that pedestrians might adapt their behavior based on the robot's approach angle. Since the investigated patterns in this present study differ in how narrow and frequent the resulting turns are, and thus lead to differences in the approaching direction of the robot, the off-set rectangular pattern could have led to stronger feelings of discomfort in the pedestrians. Further analysis of the video footage is needed to support this. Pedestrians may also perceive the off-set rectangular pattern as less predictable than the circular movement, potentially leading to more adaptations. This may also explain that the off-set rectangular pattern led to more close adaptations, as pedestrians might have misjudged the robot's next move. We also found that this effect interacted with the robot type: The larger sweeping robot following the off-set rectangular pattern resulted in fewer close adaptations, suggesting a complex interplay between robot type and movement pattern.

### *E. Strengths, Limitations, and Future Research*

The main strength of this study lies within its observation of natural human behavior during interactions with robots in a real-world environment, capturing authentic behavioral patterns and responses. It provides valuable ecological validity often missing in more controlled laboratory studies. However, several limitations should be acknowledged. First, the dataset contains an uneven distribution of undistracted and distracted pedestrians, with a low number of distracted individuals. This limits statistical power and warrants caution in interpreting findings related to distraction. Moreover, the definition of distraction was restricted to observable smartphone use, excluding other potential forms of cognitive or auditory distraction. Besides, the possibility that pedestrians disengaged from their smartphone upon encountering the robot was not accounted for. Second, clearly visible cameras may have influenced pedestrians' behavior. Third, inter-rater reliability was comparably low for some variables, which may have introduced potential biases in data interpretation. Fourth, the robots differed in multiple aspects (e.g. size, appearance, turning radius, sound, task), making it challenging to attribute behavioral differences to a single factor. Notably, the present study may also not generalize to other cultural or geographic contexts or different types of public spaces. Despite these limitations, this study provides a foundation for more controlled investigations into HRI, particularly focusing on distracted pedestrians. Future work should broaden the definition of distraction to provide a more comprehensive understanding of distraction in HRI. Future research would further benefit from implementing standardized rater training and coding protocols, as well as incorporating more objective measures such as eye-tracking (to assess the frequency and timing of distraction disengagement) and LiDAR technologies for precise movement tracking. Further, incorporating qualitative assessments of pedestrians' opinions on robot design aspects, as well as assessing their prior experience with autonomous robots, could complement behavioral observations.

### *F. Practical Implications*

The present study provides insights into the movement behaviors of distracted and undistracted pedestrians in public spaces, that can be considered when designing or choosing autonomous cleaning robots for use in public spaces. Although the number of distracted pedestrians in the present study is lower compared to prior studies, the percentage of vulnerable pedestrians is still noteworthy and should be considered when employing autonomous cleaning robots in public spaces, such as underground passages or shopping centers. The amount of displayed distraction has also been shown to vary based on the location [25], which should be considered when choosing where to deploy a robot. These findings highlight the importance of considering both robot movement patterns and physical characteristics in designing for HRI in public spaces. The research reveals that larger robots and those moving in an off-set rectangular movement pattern triggered significantly more trajectory adaptations among pedestrians. This suggests that smaller robots with more predictable, circular movement patterns might be preferable to minimize disruption in pedestrian flow. To mitigate potential conflicts, especially in areas with high distraction rates, deploying robots during less crowded times, such as nighttime, could be an effective strategy. Our research also revealed that the exact effects of movement patterns are dependent on the robot type they are deployed on, which should be kept in mind for robot deployment in public spaces. While this study did not show significant differences in movement behavior between distracted and undistracted pedestrians, considering different types of distraction in robot design remains crucial. Pedestrians wearing headphones might benefit more from visual cues of the robot, i.e. the robot projecting its path onto the floor, whereas pedestrians displaying *smombie* behavior possibly benefit more from auditive cues, such as a warning noise when the robot is approaching.

## VI. Conclusion

This field observation provides valuable insights into real HRI in public spaces, focusing on pedestrian distraction, robot type, and movement patterns. Around 8% of coded pedestrians were distracted by their phone, 29% of these not directly looking at the robot. Interestingly, distracted and undistracted pedestrians did not significantly differ in their movement behaviors around the robots, contrary to our expectations. However, the two types of robots we employed as well as their movement pattern significantly influenced lateral adaptations, with the larger sweeping robot and the off-set rectangular movement pattern eliciting more adaptations. The off-set rectangular pattern also increased close trajectory adaptations, depending on the robot type. This research contributes to understanding of how robotic presence and design affect pedestrian behavior, informing the development and deployment of more effectively integrated autonomous robots in public spaces.